\documentclass[conference]{IEEEtran}
\IEEEoverridecommandlockouts
\usepackage{cite}
\usepackage{amsmath,amssymb,amsfonts}
\usepackage{graphicx}
\usepackage{textcomp}
\usepackage{xcolor}

\usepackage{algorithm}
\usepackage{algpseudocode}

\usepackage[caption=false,font=footnotesize]{subfig}

\usepackage{graphicx}
\usepackage{tikz}
\usepackage[utf8]{inputenc}
\usepackage{pgfplots}
\DeclareUnicodeCharacter{2212}{−}
\usepgfplotslibrary{groupplots,dateplot}
\usetikzlibrary{patterns,shapes.arrows}
\pgfplotsset{compat=newest}

\usepackage{amsfonts}
\usepackage{amssymb}
\usepackage{mathrsfs}
\usepackage{amsmath}
\usepackage{mathtools}
\usepackage{url}

\usepackage{tikz}
\usepackage{pgfplots}
\DeclareUnicodeCharacter{2212}{−}
\usepgfplotslibrary{groupplots,dateplot}
\usetikzlibrary{patterns,shapes.arrows}
\pgfplotsset{compat=newest}

\usepackage{xargs}
\usepackage{color}

\definecolor{tcolor}{RGB}{0,60,104}
\usepackage{pgfplots}

\newcommand{\rvec}[1]{\ensuremath{{\boldsymbol{{#1}}}}}
\newcommand{\mat}[1]{{\ensuremath{{\mathbf{#1}}}}}

\DeclareMathOperator*{\argmax}{arg\,max}

\newcommand{\Fig}[2][KeinUnterVerweis]
{\ifthenelse{\equal{#1}{KeinUnterVerweis}}%
{Fig.~\ref{#2}}
{Fig.~\hbox{\ref{#2} #1}}%
}%

\def\BibTeX{{\rm B\kern-.05em{\sc i\kern-.025em b}\kern-.08em
    T\kern-.1667em\lower.7ex\hbox{E}\kern-.125emX}}
\begin{document}

\title{Continuous Herded Gibbs Sampling\\
\thanks{This work was funded by the Lower Saxony Ministry of Science and Culture under grant number ZN3493 within the Lower Saxony ``Vorab'' of the Volkswagen Foundation and supported by the Center for Digital Innovations (ZDIN).}
}

\author{\IEEEauthorblockN{Laura M. Wolf and Marcus Baum}
\IEEEauthorblockA{\textit{Institute of Computer Science} \\
\textit{University of G\"ottingen}\\
G\"ottingen, Germany \\
Email: \{laura.wolf, marcus.baum\}@cs.uni-goettingen.de}
}

\maketitle

\begin{abstract}
Herding is a technique to  sequentially generate deterministic samples from a probability distribution. In this work, we propose a continuous herded Gibbs sampler that combines kernel herding on continuous densities with the Gibbs sampling idea.
Our algorithm allows for deterministically sampling from high-dimensional multivariate probability densities, without directly sampling from the joint density. 
Experiments with Gaussian mixture densities indicate that the L2 error  decreases similarly to kernel herding, while the computation time is significantly lower, i.e., linear in the number of dimensions.
\end{abstract}


\section{Introduction}
Sampling, i.e., the selection of a set of points that represent a probability distribution, is an essential problem in many areas such as statistics,  machine learning, and nonlinear filtering.  
For this purpose, randomized sampling algorithms such as importance sampling \cite{sarkkaBayesianFilteringSmoothing2013} and Markov Chain Monte Carlo methods (MCMC) methods including Metropolis-Hasting  \cite{geyerPracticalMarkovChain1992}, and Gibbs Sampling \cite{gemanStochasticRelaxationGibbs1984,casellaExplainingGibbsSampler1992} are widely-used. 
  Recently, deterministic sampling methods \cite{hanebeckLocalizedCumulativeDistributions2008,frischEfficientDeterministicConditional2020} are becoming more and more relevant. Deterministically chosen samples can encode more information, thus fewer samples are required to represent a distribution \cite{chenSuperSamplesKernelHerding2010}.
Furthermore, deterministic algorithms require less testing, which is especially important in safety-critical areas such as autonomous driving \cite{koopmanChallengesAutonomousVehicle2016}.

The process of herding, introduced to generate pseudo-samples that match given moments \cite{wellingHerdingDynamicalWeights2009}, can be used to deterministically sample from discrete probability distributions. Moreover, herding has been applied to Gibbs sampling on graphical models, yielding a deterministic Gibbs sampler on discrete probability distributions \cite{chenHerdedGibbsSampling2016}. Herded Gibbs sampling also showed comparable results to randomized Gibbs Sampling considering data association in multi-object tracking \cite{wolfDeterministicGibbsSampling2020}. Bach et al. \cite{bachEquivalenceHerdingConditional2012} furthermore showed that the kernel herding procedure \cite{chenSuperSamplesKernelHerding2010} is equivalent to the Frank-Wolfe algorithm. 
By applying kernels, the herding principle was extended to continuous densities, called kernel herding  \cite{chenSuperSamplesKernelHerding2010}, yielding deterministic super-samples that contain more information than random i.i.d. samples. The herding methods will be discussed in more detail in Section~\ref{sec:background}.

Considering nonlinear filtering with deterministic samples, the unscented Kalman filter \cite{julierUnscentedFilteringNonlinear2004,wanUnscentedKalmanFilter2001} is widely-used, which employs $2d+1$ sigma-points to represent a $d$-dimensional Gaussian density. The number of samples, however, is limited and hence also the ability to represent arbitrary densities.
Deterministic Gaussian filters were created in \cite{huberGaussianFilterBased2008, steinbringS2KFSmartSampling2013} by optimizing the sample locations with respect to a suitable distance measure, where all samples are optimized simultaneously in a batch process. 
Particle filters \cite{doucetSequentialMonteCarlo2000,sarkkaBayesianFilteringSmoothing2013}, which are based on sequential Monte Carlo sampling, are often applied to nonlinear filtering problems  but they might suffer from particle degeneration such that resampling becomes necessary. In this context, particle flow filters  \cite{daumExactParticleFlow2010, dingImplementationDaumHuangExactflow2012} avoid random sampling by systematically placing the particles  based on solving partial differential equations. In \cite{lacoste-julienSequentialKernelHerding2015}, sequential Monte Carlo methods were combined with Frank-Wolfe optimization to use deterministic samples, which provided a better accuracy than random samples.

The main contribution of our paper is a new deterministic Gibbs sampling algorithm that systematically draws samples from continuous multivariate probability densities unlike the approach in \cite{chenHerdedGibbsSampling2016}, which works on discrete probability distributions and probabilistic graphical models.
For this purpose, we extend the kernel herding algorithm to Gibbs sampling, thus creating a herded Gibbs sampler for continuous densities. 
The key idea is to apply kernel herding to sample from the conditional densities in a Gibbs sampling scheme, while employing the conditional kernel density estimate in order to incorporate all previous samples into the herding process. 
Our algorithm is suited for high dimensions since only one-dimensional optimization problems have to be solved. Furthermore, it allows for generating samples from an unknown joint density, like randomized Gibbs sampling, as long as the conditional probability densities are known.
We provide experimental convergence results showing that the $L_2$ error decreases comparably to kernel herding while the computation time is significantly lower.

This paper is structured as follows:
In Section~\ref{sec:background} we revisit herding methods for deterministic sampling, including discrete herded Gibbs sampling and kernel herding. In Section~\ref{sec:chg} we then introduce our new algorithm called continuous herded Gibbs sampling and in Section~\ref{sec:evaluation} we present experimental results for Gaussian mixtures. 
The paper ends with a brief  conclusion and outlook in Section~\ref{sec:conclusion}.

\section{Background}
\label{sec:background}

In this section, we will review the basic herding \cite{wellingHerdingDynamicalWeights2009} procedure and herded Gibbs sampling \cite{chenHerdedGibbsSampling2016} for sampling from discrete distributions as well as kernel herding \cite{chenSuperSamplesKernelHerding2010} for the continuous case.

\begin{figure*}
    \centering
    \begin{tikzpicture}
    \input{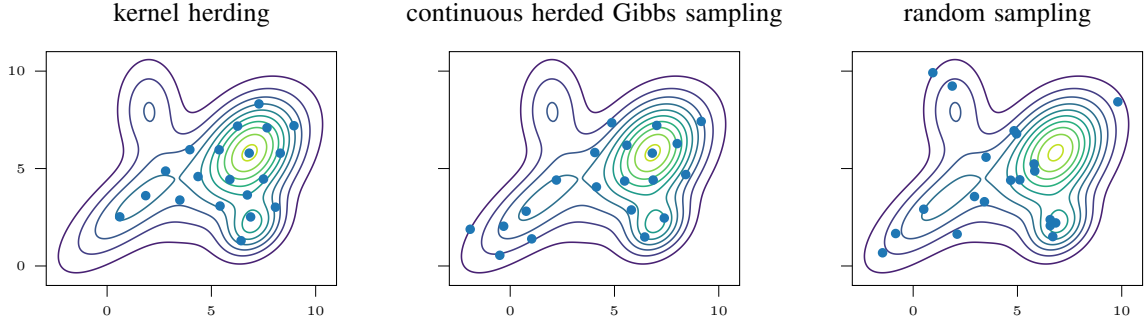}
    \end{tikzpicture}
    
    \caption{Illustration of 20 samples drawn from a Gaussian mixture with 5 components using kernel herding, continuous herded Gibbs sampling and random sampling.}
    \label{fig:example_sampling}
\end{figure*}

\subsection{Herding discrete samples}
Herding \cite{wellingHerdingDynamicalWeights2009} is a deterministic process of generating samples that match given moments $\mu = \mathbb{E}_{\mat{x}\sim P} [\phi(\mat{x})]$ with respect to the features $\phi(\mat{x})$, where $P$ is a discrete probability distribution.
The sampling is carried out via maximizing a weight vector $\mat{w}$ w.r.t. $\phi$, which includes information on all previous samples. Then, the moments and current samples are incorporated into $\mat{w}$ in an update.
The procedure to compute the $t$\textsuperscript{th} sample $\mat{x}^{(t)}\in\mathcal{X}\subseteq\mathbb{R}^d$  is given by \cite{chenHerdedGibbsSampling2016}
\begin{align}
    \mat{x}^{(t)} &= \argmax_{\mat{x}\in \mathcal{X}}\left\langle \mat{w}^{(t-1)},\rvec{\phi}(\mat{x})\right\rangle, \label{eq:herding_max}\\
    \mat{w}^{(t)} &= \mat{w}^{(t-1)} + \mu - \rvec{\phi}\left(\mat{x}^{(t)}\right), \label{eq:herding_update}
\end{align}
with the weight vector $\mat{w}\in\mathcal{H}$ and feature map $\rvec{\phi}:\mathcal{X}\mapsto\mathcal{H}$. The feature space $\mathcal{H}$ is a Hilbert space with inner product $\langle\cdot,\cdot\rangle$. In order to sample from a probability distribution, i.e. $\mu=P$, the indicator vector is used as feature map $\rvec{\phi}$. The weight vector $\mat{w}$ can be initialized as $\mu$. With $\phi$ as the indicator vector, \eqref{eq:herding_max} boils down to maximizing $\mat{w}$, \eqref{eq:herding_update} results in a subtraction of $1$ at the corresponding element of $\mat{w}$. The probability of $\mat{x}$ determines how fast it will be sampled again, as $P$ is added to $\mat{w}$ after every sample. 

In \cite{chenHerdedGibbsSampling2016}, the same process was applied to the full conditionals in Gibbs sampling, thus creating a deterministic Gibbs sampler for discrete probability distributions. However, herding has to be applied separately to each conditional probability distribution, maintaining separate weight vectors for all conditionals and resulting in a worst case exponential space complexity. Pseudocode generating $T$ samples using herded Gibbs sampling can be found in Alg.~\ref{alg:herdedgibbs}. We make use of the notation $\mat{\bar{x}}_i = (\mat{x}_1,\dots,\mat{x}_{i-1},\mat{x}_{i+1},\dots,\mat{x}_d)^\top$. The weights $\mat{w}^{i,\mat{\bar{x}}_i}$ are initialized with $P(\mat{x}_i|\mat{\bar{x}}_i)$. Note, that there is one weight vector for each combination of $i$ and $\mat{\bar{x}}_i$. 

\begin{algorithm}
\caption{Discrete herded Gibbs sampling \cite{chenHerdedGibbsSampling2016}}
\label{alg:herdedgibbs}
\begin{algorithmic}[1]
    \For{$t=1, \dots, T$}
        \For{$i=1, \dots, d$}
            \State $\mat{x}_i = \argmax \mat{w}^{i,\bar{\mat{x}}_i}$
            \State $\mat{w}^{i,\bar{\mat{x}}_i} = \mat{w}^{i,\bar{\mat{x}}_i} + P\left(\mat{x}_i \mid  \mat{\bar{x}}_i \right)$ 
            \State $\mat{w}^{i,\bar{\mat{x}}_i}_{\mat{x}_i} = \mat{w}^{i,\bar{\mat{x}}_i}_{\mat{x}_i} -1$
        \EndFor
        \State $\mat{x}^{(t)}=\mat{x}$
    \EndFor
\end{algorithmic}
\end{algorithm}

\subsection{Kernel herding}
The herding process was extended to continuous probability densities in \cite{chenSuperSamplesKernelHerding2010}.  The objective is to deterministically determine $t$ samples $\mat{x}^{(1)},\dots,\mat{x}^{(t)}\in\mathbb{R}^d$ that represent a $d$-dimensional  probability density $p$.
In the continuous case of herding, an infinite number of features needs to be considered, which is intractable. 
By applying kernels $k(\mat{x},\mat{x'}) = \langle \phi(\mat{x}),\phi(\mat{x}')\rangle$, the feature vectors are incorporated implicitly. A weight vector is not needed anymore, it is replaced by a weight function, which incorporates the expectation of the kernel over $p$, the moments to be matched, as well as all previous samples and needs to be optimized. Sample $\mat{x}^{(t+1)}$ from a probability density $p(\mat{x})$ is then computed as follows, where both the sampling step and the weight update are combined into one optimization  \cite{chenSuperSamplesKernelHerding2010}:
\begin{align}
  \mat{x}^{(t+1)} = \argmax_{\mat{x}\in\mathcal{X}} \mathbb{E}_{\mat{x}'\sim p}\left[k(\mat{x},\mat{x}')\right] 
  - \frac{1}{t+1}\sum_{s=1}^t k\left(\mat{x},\mat{x}^{(s)}\right).
  \label{eq:KernelHerding}
\end{align}

\section{Continuous herded Gibbs sampling}
\label{sec:chg}
In this work, we propose  to combine herded Gibbs sampling with kernel herding to create a deterministic Gibbs sampler for continuous distributions. A naïve attempt would be to apply kernel herding directly to the conditional probabilities of Gibbs sampling. In \eqref{eq:KernelHerding} one would replace $\mat{x}^{(t+1)}$ by $\mat{x}^{(t+1)}_i$ and $p(\mat{x})$ by $p(\mat{x}_i|\mat{\bar{x}}_i)$ in \eqref{eq:KernelHerding}, where $\mat{\bar{x}}_i$ is defined as above.
In the summation in \eqref{eq:KernelHerding} only those samples $\mat{x}^{(s)}_i$ would be used, where $\mat{\bar{x}}_i^{(s)}=\mat{\bar{x}}_i$. This would be equivalent to having a separate weight vector for each combination of $i$ and $\mat{\bar{x}}_i$, as it is the case with discrete herded Gibbs sampling. However, since the number of possible $\mat{\bar{x}}_i$'s is uncountable in the continuous case, a second sample with $\mat{\bar{x}}_i^{(s')}=\mat{\bar{x}}_i^{(s)}$ will very unlikely appear again. This would result in starting a new herding process for almost every sample, which is undesirable, as the valuable information of previous samples is not used. 

For this reason, our key idea is to include all previous samples into the herding process by replacing the kernel density estimate in kernel herding with the conditional kernel density estimate. For this purpose, we rewrite \eqref{eq:KernelHerding} as
\begin{equation}
    \mat{x}^{(t+1)} = \argmax_{\mat{x}\in\mathcal{X}} \frac{t+1}{t}\mathbb{E}_{\mat{x}'\sim p}\left[k(\mat{x},\mat{x}')\right] - \tilde{p}_t(\mat{x}),  
\end{equation}
with the kernel density estimate 
\begin{equation}
    \tilde{p}_t(\mat{x}) = \frac{1}{t} \sum_{s=1}^t k\left(\mat{x},\mat{x}^{(s)}\right).
\end{equation}

In order to sample from the conditional probability density, we consider the kernel density estimate conditioned on $\mat{\bar{x}}_i$
\begin{equation}
    \tilde{p}_t(\mat{x}_i|\mat{\bar{x}}_i) = \frac{\tilde{p}_t(\mat{x})}{\tilde{p}_t(\mat{\bar{x}}_i)}
    = \frac{\sum_{s=1}^t k\left(\mat{x},\mat{x}^{(s)}\right)}{\sum_{s=1}^t k\left(\mat{\bar{x}}_i,\mat{\bar{x}}_i^{(s)}\right)}.
\end{equation}

If the dimensions of the kernel are independent, as it is the case, e.g., for Gaussian kernels, we can rewrite the conditional kernel density estimate as 
\begin{align}
    \tilde{p}_t(\mat{x}_i|\mat{\bar{x}}_i) &= \frac{\sum_{s=1}^t k\left(\mat{x}_i,\mat{x}_i^{(s)}\right)k\left(\mat{\bar{x}}_i,\mat{\bar{x}}_i^{(s)}\right)}{\sum_{s=1}^t k\left(\mat{\bar{x}}_i,\mat{\bar{x}}_i^{(s)}\right)} \\
    &= \frac{1}{c}\sum_{s=1}^t w_s k\left(\mat{x}_i,\mat{x}_i^{(s)}\right),
\end{align}
with $w_s = k\left(\mat{\bar{x}}_i,\mat{\bar{x}}_i^{(s)}\right)$ and $c= \sum_{s=1}^t w_s$.

We can then formulate the procedure for continuous herded Gibbs sampling as follows
\begin{align}
    \mat{x}_i^{(t+1)} = &\argmax_{\mat{x}_i\in\mathbb{R}} \frac{t+1}{t} \mathbb{E}_{x'\sim p(\mat{x}_i|\mat{\bar{x}}_i)}[k(\mat{x}_i,x')]\notag\\
    &\,- \tilde{p}_t(\mat{x}_i|\mat{\bar{x}}_i).
\end{align}
With independent kernels, this simplifies to
\begin{align}
    \mat{x}_i^{(t+1)} = &\argmax_{\mat{x}_i\in\mathbb{R}} \frac{t+1}{t} \mathbb{E}_{x'\sim p(\mat{x}_i|\mat{\bar{x}}_i)}[k(\mat{x}_i,x')]\notag\\
    &\,- \frac{1}{c}\sum_{s=1}^t w_s k\left(\mat{x}_i,\mat{x}_i^{(s)}\right).
    \label{eq:ContHGsimple}
\end{align}
\eqref{eq:ContHGsimple} shows that all previous samples are incorporated into the herding process, however, they are weighted according to the distance between $\mat{\bar{x}}_i$ and $\mat{\bar{x}}_i^{(s)}$. Pseudocode for continuous herded Gibbs sampling can be found in Alg.~\ref{alg:contHG}.
Fig.~\ref{fig:example_sampling} shows samples drawn from a Gaussian mixture using kernel herding, continuous herded Gibbs sampling and random sampling from the joint density.

\begin{algorithm*}
\begin{algorithmic}
    \For{$t=1, \dots, T$}
        \For{$i=1, \dots, d$}
            \State $\mat{x}_i = \argmax_{x_i\in\mathbb{R}} \frac{t+1}{t} \mathbb{E}_{x'\sim p(\mat{x}_i|\mat{\bar{x}}_i)}[k(\mat{x}_i,x')] 
    - \frac{1}{c}\sum_{s=1}^t w_s k\left(\mat{x}_i,\mat{x}_i^{(s)}\right)$
        \EndFor
        \State $\mat{x}^{(t)}=\mat{x}$
    \EndFor
\end{algorithmic}
\caption{Continuous herded Gibbs sampling}
\label{alg:contHG}
\end{algorithm*}

\begin{figure}
    \centering
    \begin{tikzpicture}[scale=0.8]
    \input{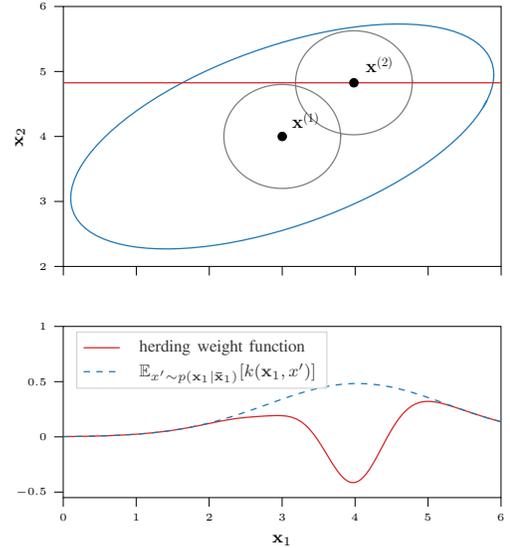}
    \end{tikzpicture}
    
    \caption{Visualization of continuous herded Gibbs sampling. Upper plot: two samples drawn from a Gaussian. The kernels and Gaussian are visualized with their $2\sigma$-bound. The red line depicts $\mat{\bar{x}}_1$. The lower plot shows the herding weight function \eqref{eq:ContHGsimple} (solid red line) and $\mathbb{E}_{x'\sim p(\mat{x}_1|\mat{\bar{x}}_1)}[k(\mat{x}_1,x')]$ (dashed blue line).}
    \label{fig:example_chg}
\end{figure}

A visualization of our algorithm is provided in Fig.~\ref{fig:example_chg}. It shows two samples, $\mat{x}^{(1)}$ and $\mat{x}^{(2)}$ drawn from a Gaussian density using Gaussian kernels. In the next step $\mat{x}_1^{(3)}$ will be sampled along the red line in the upper plot, which depicts $\mat{\bar{x}}_1$. 
The lower plot shows the expectation $\mathbb{E}_{x'\sim p(\mat{x}_1|\mat{\bar{x}}_1)}[k(\mat{x}_1,x')]$ (dashed blue line) and the herding weight function (solid red line), see \eqref{eq:ContHGsimple}. Note that the difference between both functions is given by $\frac{1}{c}\sum_{s=1}^t w_s k\left(\mat{x}_1,\mat{x}_1^{(s)}\right)$ in the Gaussian case and both functions are equal before any samples are drawn. The impact of a previous sample $\mat{x}^{(s)}$ on the herding weight function is determined by $w_s$, which resembles the distance of $\mat{x}^{(s)}$ to $\mat{\bar{x}}_1$ (solid red line in the upper plot). This can also be seen in Fig.~\ref{fig:example_chg}, where $\mat{x}^{(2)}$ has a higher impact on the weighting function with $w_2=0.997$ as opposed to $\mat{x}^{(1)}$ with $w_1=0.118$, which has a higher distance to $\mat{\bar{x}}_1$. In order to determine $\mat{x}_1^{(3)}$, the herding weight function is maximized. Afterwards, the process is repeated for the second dimension.

\section{Evaluation on Gaussian mixtures}
\label{sec:evaluation}
In order to demonstrate the performance of the proposed continuous herded Gibbs sampling algorithm, we provide numerical results for the problem of  sampling from  Gaussian mixture densities. A Gaussian mixture with $M$ components has the density function $p(\mat{x})=\sum_{i=1}^M \phi_i p_i(\mat{x})$, with $p_i(\mat{x})\sim \mathcal{N}(\rvec{\mu}_i,\Sigma_i)$, $\phi_i\geq 0$ and $\sum_{i=1}^M \phi_i=1$. We employed a Gaussian kernel
\begin{equation}
    k_d(\mat{x},\mat{x}') = \frac{1}{\sqrt{ (2\pi)^d|\bar{\Sigma}|}} e^ {-\frac{1}{2}(\mat{x}-\mat{x}')^\top\bar{\Sigma}^{-1}(\mat{x}-\mat{x}')}   
\end{equation}
with standard deviation $\sigma_k$ and $\bar{\Sigma} = \sigma_k^2\mat{I}_d$. 

The expectation in \eqref{eq:KernelHerding} can then be calculated as 
\begin{align}
  &\mathbb{E}_{\mat{x}'\sim p}\left[k_d(\mat{x},\mat{x}')\right] =  \int_{\mathbb{R}^d} k_d(\mat{x},\mat{x}')p(\mat{x}')\mathrm{d}\mat{x}'  \notag\\
  &\quad= \sum_{i=1}^M \phi_i \frac{1}{(2\pi)^d \left| \Sigma_i + \bar{\Sigma} \right|}
  e^{-\frac{1}{2} (\mat{x}-\rvec{\mu}_i)^\top (\Sigma_i+\bar{\Sigma})^{-1}(\mat{x}-\rvec{\mu}_i)},
  \label{eq:expectation_herding}
\end{align}
using \cite{jebaraBhattacharyyaExpectedLikelihood2003} and the identity $(\mat{A}^{-1} + \mat{B}^{-1})^{-1} = \mat{A}(\mat{A}+\mat{B})^{-1}\mat{B} = \mat{B}(\mat{A}+\mat{B})^{-1}\mat{A}$.
Since the conditional of a Gaussian mixture is also a Gaussian mixture, \eqref{eq:expectation_herding} can also be applied for the expectation in \eqref{eq:ContHGsimple}, which is a one-dimensional function. 

\begin{figure*}[p]
    
    \centering
    \subfloat[Two-dimensional space.\label{fig:error_2d}]{
        \begin{tikzpicture}
        \input{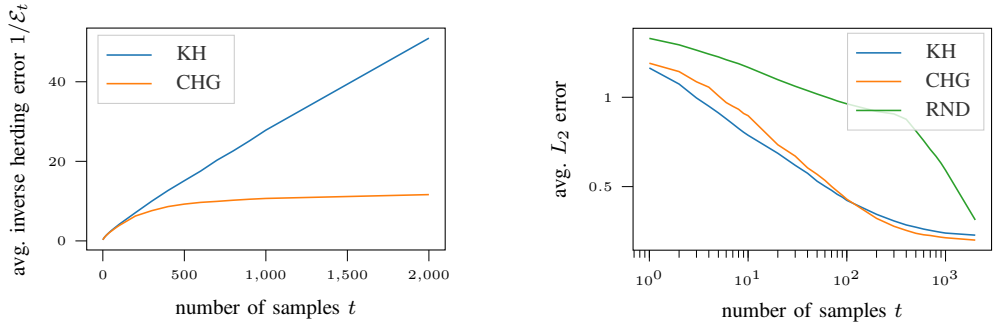}
        \end{tikzpicture}
    }
    \\
    \subfloat[Ten-dimensional space.\label{fig:error_10d}]{
        \begin{tikzpicture}
        \input{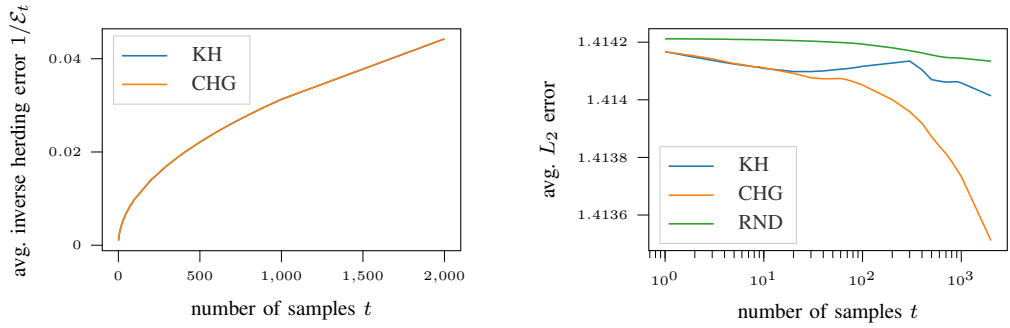}
        \end{tikzpicture}
    }
    
    \caption{Inverse herding error $1/\varepsilon_t$ and $L_2$ error for kernel herding (KH) and continuous herded Gibbs sampling (CHG) averaged over 10 random Gaussian mixtures each. The $L_2$ error is also computed for random samples (RND) drawn from the joint distribution. }
    \label{fig:herding_error}
\end{figure*}

\begin{figure*}[p]
    
    \centering
    \subfloat[\label{fig:error_example_good}Continuous herded Gibbs converged in both the herding error and the $L_2$ error.]{
        \begin{tikzpicture}[scale=0.95]
        \input{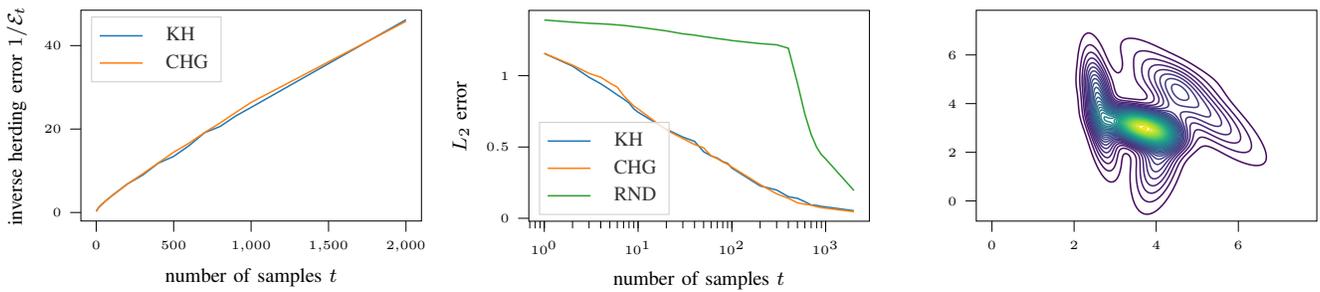}
        \end{tikzpicture}
    }
    \\
    \subfloat[\label{fig:error_example_bad}Continuous herded Gibbs only converged in the $L_2$ error.]{
        \begin{tikzpicture}[scale=0.95]
        \input{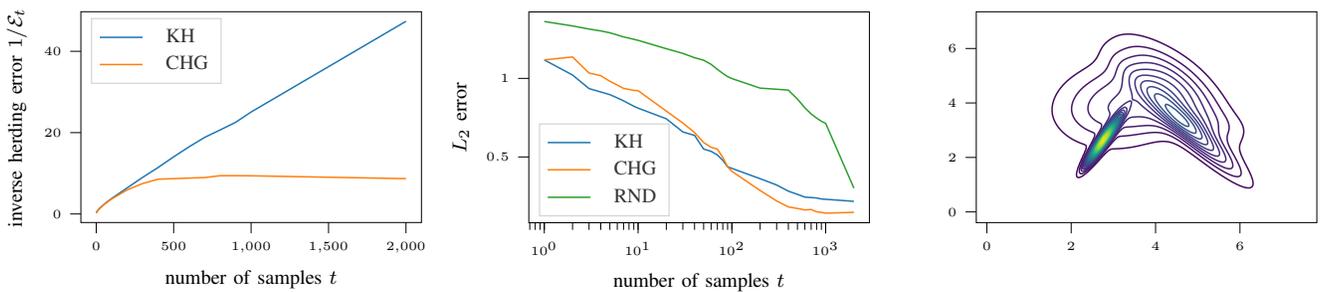}
        \end{tikzpicture}
    }
    
    \caption{Two example Gaussian mixtures in two-dimensional space, with errors as in Fig.~\ref{fig:herding_error} and contour plot of the corresponding Gaussian mixture.}
    \label{fig:error_example}
\end{figure*}

We used a heuristic approach to determine the starting points for the optimization, based on component means and covariances of the Gaussian mixtures. 
This results in a higher number of starting points for kernel herding in higher dimensions, which, however, is needed to avoid local minima. For continuous herded Gibbs sampling the number of starting points does not depend on the dimension, as the weight function is one-dimensionl, but the number of optimizations per sample increases nonetheless. 
We used the BFGS algorithm \cite{nocedalQuasiNewtonMethods2006} for optimization with default parameters as provided by scipy for kernel herding as well as continuous herded Gibbs sampling. 
The first sample of the continuous herded Gibbs sampler was initialized as the first sample generated by kernel herding\footnote{Source code is available at \url{https://github.com/Fusion-Goettingen.}}.

First, we computed the herding error $\mathcal{E}_t$ \cite[eq. (9)]{chenSuperSamplesKernelHerding2010}, which measures the distance between $p$ and the set of samples in the feature space induced by the applied kernel, i.e.,
\begin{align}
\label{eq:herding_error}
    \mathcal{E}^2_t = & \left|\left|\mathbb{E}_{x\sim p} [\phi(x)] - \frac{1}{t}\sum_{s=1}^t\rvec{\phi}\left(\mat{x}^{(s)}\right)\right|\right|^2  \notag \\
    =& \:\mathbb{E}_{\mat{x},\mat{x}'\sim p}\left[k(\mat{x},\mat{x}')\right] 
-\frac{2}{t}\sum_{s=1}^t \mathbb{E}_{\mat{x}\sim p}\left[k\left(\mat{x},\mat{x}^{(s)}\right)\right]  \notag\\
&+\frac{1}{t^2}\sum_{s,s'=1}^t k\left(\mat{x}^{(s)},\mat{x}^{s'}\right)  .
\end{align}
In the Gaussian case the first expectation in \eqref{eq:herding_error} is calculated analogous to \eqref{eq:expectation_herding}:
\begin{align}
  &\mathbb{E}_{\mat{x},\mat{x}'\sim p}\left[k_d(\mat{x},\mat{x}')\right] =  \int_{\mathbb{R}^d} k_d(\mat{x},\mat{x}')p(\mat{x})p(\mat{x}')\mathrm{d}\mat{x}\mathrm{d}\mat{x}'  \notag\\
  &\quad= \sum_{i,j=1}^M \phi_i \phi_j \frac{1}{(2\pi)^d \left| \Sigma_i + \Sigma_j + \bar{\Sigma} \right|} \;\times \notag \\
  &\qquad \;\; e^{-\frac{1}{2} (\rvec{\mu}_i - \rvec{\mu}_j)^\top (\Sigma_i+\Sigma_j+\bar{\Sigma})^{-1}(\rvec{\mu}_i - \rvec{\mu}_j)}.
  \label{eq:expectation_herding_error}
\end{align}

The herding error is minimized greedily by kernel herding~\cite{chenSuperSamplesKernelHerding2010}. We included it in order to determine if continuous herded Gibbs sampling can generate similar samples as kernel herding, even though it employs a different objective function. 

We computed $\mathcal{E}_t$ on 10 randomly constructed two-dimensional and ten-dimensional Gaussian mixtures with 5 components. The samples were generated using a Gaussian kernel with $\sigma_k=0.1$.  
Fig.~\ref{fig:error_2d} shows that $1/\mathcal{E}_t$ increases linearly with $t$ for kernel herding in the two-dimensional example, as was also found in \cite{chenSuperSamplesKernelHerding2010}. For continuous herded Gibbs sampling it increases sublinearly, indicating that continuous herded Gibbs sampling does not converge in the feature space of kernel herding. It has to be noted though, that these results highly depend on the chosen distribution. In some cases the convergence of continuous herded Gibbs sampling is as good as kernel herding (Fig.~\ref{fig:error_example_good}), whereas on others the herding error stagnates (Fig.~\ref{fig:error_example_bad}). The Gaussian mixture in Fig.~\ref{fig:error_example_bad} has a very narrow and long component as opposed to the Gaussian mixture in Fig.~\ref{fig:error_example_good}. This could lead to problems with our approach, if the kernel size is too big, as each dimension is sampled individually. A solution to this problem might be adapting the kernel size based on the covariances' eigenvalues. In the ten-dimensional example (Fig.~\ref{fig:error_10d}) both algorithms showed the same slow convergence in the herding error. This could be due to a too small kernel size for higher dimensions.

Second, we calculated the normalized $L_2$ distance as described in \cite{jensenEvaluationDistanceMeasures2007}, which calculates the distance in function space between the original Gaussian mixture density and the kernel density estimate that arises from the combination of the samples  with the Gaussian kernel. 
In contrast to the herding error, the $L_2$ error does not incorporate the kernels into the error computation, they are only applied to construct a density of the samples. 
That way both herding variants can also be compared with random samples, where we averaged over 20~Monte Carlo runs. The random samples where directly drawn from the joint distribution. We used the same densities as above. Fig.~\ref{fig:error_2d} shows that continuous herded Gibbs sampling and kernel herding yield comparable results in two-dimensional space, with kernel herding yielding a better performance with few samples (less than 100) and continuous herded Gibbs sampling yielding a better performance with more than 100 samples. However, both seem to converge in the function space and show better results than random sampling.  This shows that both herding variants are capable of representing a Gaussian mixture with fewer samples than random sampling. It has to be noted, that continuous herded Gibbs sampling even converges in the $L_2$ distance on the example in Fig.~\ref{fig:error_example_bad}, where the herding error did not converge. In ten-dimensional space (Fig.~\ref{fig:error_10d}), continuous herded Gibbs converges much faster than kernel herding, which almost does not converge just as random sampling. Furthermore, the $L_2$ error only slightly decreases. This could be caused by an inappropriate kernel size as well as a too small sample size for higher dimensions.

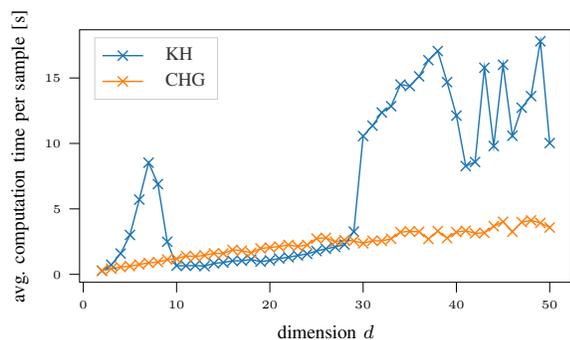
\begin{figure}
    \centering
    \begin{tikzpicture}[scale=0.9]
\definecolor{color0}{rgb}{0.12156862745098,0.466666666666667,0.705882352941177}
\definecolor{color1}{rgb}{1,0.498039215686275,0.0549019607843137}

\begin{axis}[
height=0.6\linewidth,
label style={font=\footnotesize},
legend cell align={left},
legend style={
  fill opacity=0.8,
  draw opacity=1,
  text opacity=1,
  at={(0.03,0.97)},
  anchor=north west,
  draw=white!80!black
},
legend style={font=\footnotesize},
tick align=outside,
tick label style={font=\tiny},
tick pos=left,
width=\linewidth,
x grid style={white!69.0196078431373!black},
xlabel={dimension \(\displaystyle d\)},
xmin=-0.4, xmax=52.4,
xtick style={color=black},
y grid style={white!69.0196078431373!black},
ylabel={avg. computation time per sample [s]},
ymin=-0.60677225, ymax=18.67737725,
ytick style={color=black}
]
\addplot [semithick, color0, mark=x, mark size=3, mark options={solid}]
table {%
2 0.26978
3 0.726533
4 1.586857
5 3.004922
6 5.709458
7 8.528176
8 6.89664
9 2.488196
10 0.659618
11 0.652425
12 0.675979
13 0.626131
14 0.838338
15 0.901607
16 1.030906
17 1.039189
18 1.139937
19 0.967602
20 1.067855
21 1.210828
22 1.308743
23 1.440967
24 1.558992
25 1.781063
26 1.948965
27 2.088359
28 2.281161
29 3.264003
30 10.559444
31 11.354806
32 12.360472
33 12.841875
34 14.495293
35 14.379976
36 15.124127
37 16.355465
38 17.058803
39 14.681619
40 12.121521
41 8.269231
42 8.581831
43 15.776615
44 9.802179
45 16.007755
46 10.580122
47 12.72454
48 13.609098
49 17.800825
50 10.02846
};
\addlegendentry{KH}
\addplot [semithick, color1, mark=x, mark size=3, mark options={solid}]
table {%
2 0.282067
3 0.430161
4 0.55901
5 0.623143
6 0.755222
7 0.874868
8 0.935277
9 1.131199
10 1.187671
11 1.404524
12 1.361096
13 1.453966
14 1.601653
15 1.593787
16 1.865752
17 1.806538
18 1.644563
19 1.976137
20 2.050869
21 2.124626
22 2.245603
23 2.132617
24 2.248792
25 2.732489
26 2.788224
27 2.48941
28 2.616799
29 2.554677
30 2.371904
31 2.568743
32 2.557246
33 2.714641
34 3.224669
35 3.282892
36 3.2502
37 2.700429
38 3.304678
39 2.770472
40 3.221878
41 3.328548
42 3.133281
43 3.173539
44 3.675802
45 4.016586
46 3.258685
47 3.991953
48 4.12255
49 3.927926
50 3.57827
};
\addlegendentry{CHG}
\end{axis}
    \end{tikzpicture}
    \caption{Computation time for kernel herding and continuous herded Gibbs sampling.}
    \label{fig:run_time}
\end{figure}

In order to compare the computation time, we randomly generated 10 Gaussian mixtures with 5 components and computed 100 samples with each method (kernel size $\sigma_k = 0.1$). We repeated this process for Gaussian mixtures from dimension 2 to 50. The results depicted in Fig.~\ref{fig:run_time} show that the average computation time of continuous herded Gibbs sampling increases linearly in the number of dimensions, while the average computation time for kernel herding is much higher below 10 and above 30 dimensions, in between it is a little bit faster than continuous herded Gibbs sampling.


\section{Conclusion and future work}\label{sec:conclusion}

In this paper, we proposed the continuous herded Gibbs sampler, a deterministic sampling algorithm for multivariate probability densities. As only one-dimensional optimization problems have to be solved, it is suited for high dimensional problems. The experiments showed that continuous herded Gibbs sampling can converge as fast as kernel herding, however, the performance in terms of the herding error heavily depends on the chosen Gaussian mixture although it always converges in the $L_2$ distance in our experiments. 

Future research should include further investigating the convergence of the algorithm including strategies to find the optimal kernel size for a density. The results regarding the computation time indicate that continuous herded Gibbs sampling is also promising for high-dimensional applications. It should be investigated if the computation time can be optimized, e.g., by choosing a fixed number of random starting points for the optimization, which could be relevant for Gaussian mixtures with many components. Furthermore, the application to other families of distributions and kernels should be topic to future research.




\bibliographystyle{IEEEtran}
\bibliography{HerdedGibbsContinuous}

\end{document}